# Healthcare Knowledge Graph Construction: State-of-the-art, open issues, and opportunities

Bilal Abu-Salih[1], Muhammad AL-Qurishi[2], Mohammed Alweshah[3], Mohammad AL-Smadi[4,5],

Reem Alfayez[1], Heba Saadeh[1],

[1] The University of Jordan, Jordan
[2] King Saud University, Saudi Arabia
[3] Al-Balqa Applied University
[4] Jordan University of Science and Technology
[5] Qatar University

## Abstract

The incorporation of data analytics in the healthcare industry has made significant progress, driven by the demand for efficient and effective big data analytics solutions. Knowledge graphs (KGs) have proven utility in this arena and are rooted in a number of healthcare applications to furnish better data representation and knowledge inference. However, in conjunction with a lack of a representative KG construction taxonomy, several existing approaches in this designated domain are inadequate and inferior. This paper is the first to provide a comprehensive taxonomy and a bird's eye view of healthcare KG construction. Additionally, a thorough examination of the current state-of-the-art techniques drawn from academic works relevant to various healthcare contexts is carried out. These techniques are critically evaluated in terms of methods used for knowledge extraction, types of the knowledge base and sources, and the incorporated evaluation protocols. Finally, several research findings and existing issues in the literature are reported and discussed, opening horizons for future research in this vibrant area.

***Keywords*:** Knowledge Graph; Knowledge Graph Construction; Healthcare Knowledge Graph; drugs; diseases; biomedicine; survey.

## 1. Introduction

The emergence of big data has opened up new possibilities and ushered in significant changes in various disciplines. Healthcare industry is one of such areas in which advanced and sophisticated data analysis is required to accommodate and properly understand the growing volume of healthcare data, thereby optimising healthcare delivery. However, healthcare data is still regarded as a by-product [1], thus massive healthcare data sources remain neglected and underutilised [1, 2]. Attaining meaningful and actionable knowledge from such data sources could positively affect patient care and enable more accurate diagnosis, prevention of disease, personalised treatment, and better decision-making. Primary obstacles for analysts include heterogeneity of healthcare data sources and formats, lexical disparities, and the lack of comprehensive and integrated healthcare knowledge libraries [3].

Knowledge Graphs (KGs) have evolved into a new type of knowledge representation that serves as the cornerstone for a variety of applications ranging from general to specialised industrial use [4, 5]. The fundamentally abstract structure of this technology, which efficiently promotes domain conceptualisation and data management, is one of the key factors driving the growing interest in it. The KG, in particular, displays an integrated collection of real-world entities linked by semantically associated relationships. In this case, data annotation put the available semantic content in a machine-readable format, minimising ambiguity and generating relevant information particular to the domain of an application. KGs can furnish an efficient and effective technical solution to conceptualise a healthcare domain and thus be used for several downstream tasks. Therefore, incorporating this technology into healthcare data analytics has emerged as a solution capable to mitigate such issues as data island's complexity, heterogeneity, and sheer size. However, constructing healthcare KGs with unproven methodologies raises concerns regarding their quality and robustness and whether sufficient assessment measures have been applied, especially for KGs obtained from unstructured data sources (such as scientific medical literature or social media). Furthermore, the dynamic nature of healthcare data is strongly linked to context, and numerous facts that characterise clinical and medical entities may vary or change over time. Disregarding the flexibility of knowledge lowers the quality and accuracy of facts embedded in the KGs, thereby leading to substandard decision-making based only on such data sources. As a result, it is critical to perform a detailed analysis of current state-of-the-art methodologies for healthcare KG creation in order to identify such difficulties and open new avenues for pursuing potential solutions.

This survey offers a bird's eye view of the current construction techniques and possible applications of KG technology in the healthcare domain. First, a taxonomy of healthcare KG construction is formulated to illustrate the scope of usage of KG in healthcare, levels of knowledge extraction, different types of knowledge bases and sources, and existing evaluation procedures. Next, we examined significant state-of-the-art KG generation approaches relevant for critical healthcare applications, including (i) drug discovery, repurposing and adverse reactions; (ii) diseases and disorders; (iii) biomedicine; and (iv) other miscellaneous healthcare applications. These approaches are scrutinized, with a summary created for each domain demonstrating specific KG functionalities, the incorporated knowledge extraction techniques (at both entity and relation levels), type of the knowledge base, the resources needed to construct it, relevant KG statistics, the measurements used to assess the KG construction methodology, and the limitations and shortcomings of each approach. This paper is distinguished from similar works that tend to focus too narrowly on specific healthcare subdomains [6, 7] or generic applications of KG in healthcare [8]. In particular, the following are the key contributions of this paper:

- To the best of our knowledge, this survey is the first to provide a bird's eye view of healthcare KG construction.
- A new representative taxonomy is outlined to facilitate easier KG construction in the healthcare domain.
- An in-depth analysis of state-of-the-art KG construction methodologies is provided, and their main strengths and weaknesses are discussed.
- A summary of the research findings and remaining issues is presented, paving the way for future research.

In Section 2, taxonomy of KG construction in healthcare is presented and analyzed from multiple perspectives. Several KG construction approaches relevant for various healthcare domains are reported in Section 4. Section 5 summarizes the major flows of the existing techniques, and the observed research gaps, and offers suggestions to overcome them.

## 2. Survey Methodology

This paper aims to review the recent KG construction approaches for healthcare applications. Thus, we attempt to cover all papers that describe mechanisms for KG construction to benefit the healthcare domain. We focus on articles that were published in the past five years (2018-2022). PRISMA (Preferred Reporting Items for Systematic Reviews and Meta-Analyses) framework [9] is followed to guide this systematic review. As demonstrated in Figure 1, around 560 articles were selected in the first stage from various databases including Elsevier, ACM Digital Library, Multidisciplinary Digital Publishing Institute (MDPI), IEEE Xplore digital library, and Google Scholar. The collected articles were all in English and were retrieved using the following keywords used in this query: "Knowledge Graph Construction", "Healthcare", "biomedicine", "medicine", "drug discovery", "drug repurposing", "adverse drug reaction", "disease(s)", "disorder", etc. An additional 83 articles were identified and added to the corpus by reviewing the citations map of the tentative collected set of papers. The first stage resulted in a total of 643 records. Another round of inspection was carried out in the screening stage to eliminate any redundant or irrelevant articles. This was accomplished by examining both the title and the abstract of each paper. In this way, 440 records were excluded in the screening stage as they did not meet the inclusion criteria. In particular, many of the articles discussed approaches for KG embeddings that are applied to existing KGs, thus no construction of new healthcare KGs was proposed. Another array of articles reported KG construction for other domains of knowledge yet indicated "healthcare" as an example of the popularity of KGs to tackle industrial applications. The eligibility phase was then carried out by examining the full text of papers and eliminating the irrelevant ones (102 records). In the final stage, a total of 101 papers were deemed to be qualified to be included in this review.

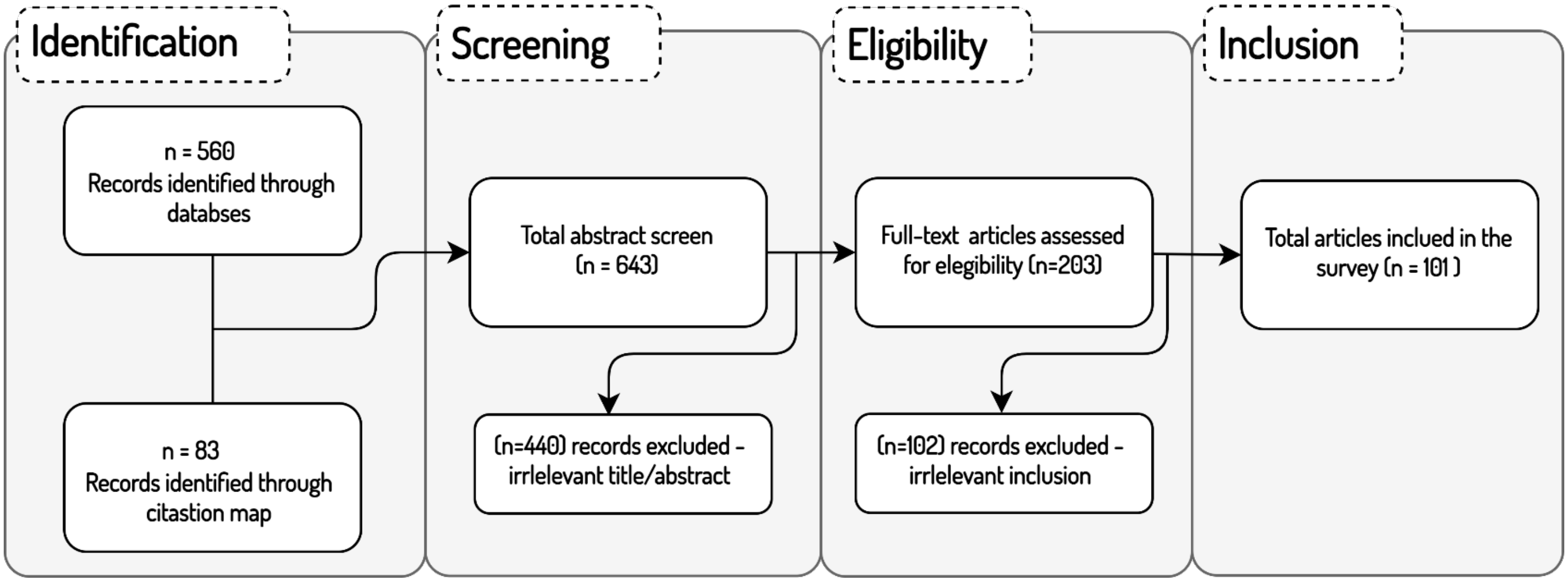

*Figure 1: The article selection strategy for the literature review (PRISMA model).*

Figure 2 presents the volume distribution of the selected articles over the past years, clearly showing the growing interest in this technology.

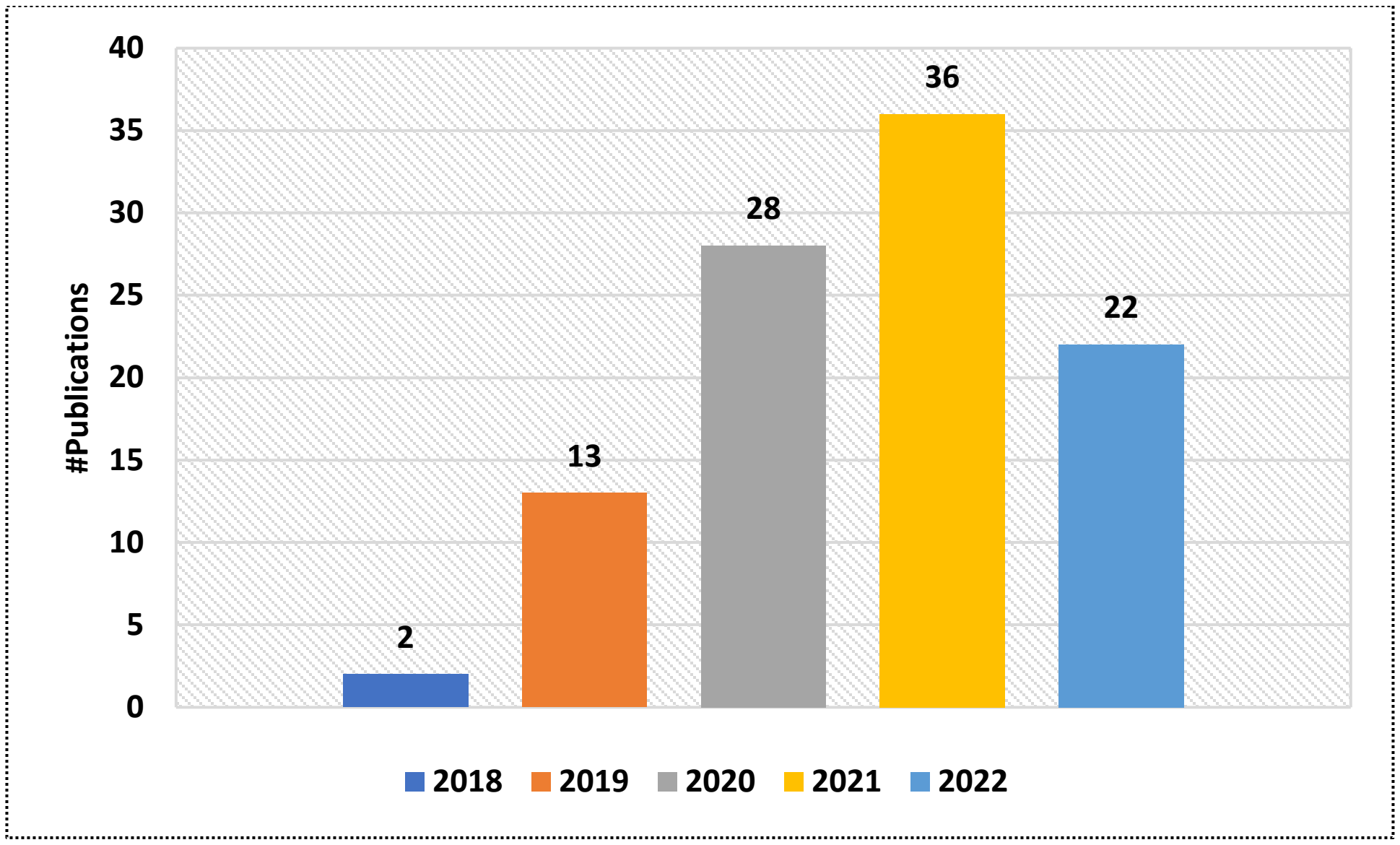

*Figure 2: #Publications about KG construction for healthcare in the past years.*

# 3. Groundworks

## 3.1 An overview of KG

A KG is a multidimensional graph that contains entities (nodes) and relations (edges) that describe the interrelation of one or more domains. Hence, the KG displays a unified collection of real-world objects connected by semantically relevant relationships. The concept of semantic interlinking is framed by Semantic Web technology whereby data can be annotated in a machine-interpretable format. This is commonly accomplished through the use of ontologies, which define concepts (representing a collection of entities), the relations between entities, and semantic rules, thereby giving a formal and explicit representation of that domain's knowledge [10, 11]. These efforts are fostered by using KGs, an abstract data model that captures a single standard representation of semantically related data (i.e., a graph).

A KG is a directed graph ($G$), where $G = (V, E)$. This notation depicts the relationship between entities, as well as the interactions between these entities, in terms of graph vertices ($V$) and edges ($E$) connecting these vertices. The edges reflect relationships between real-world things whereas the vertices represent real-world entities. The edges of the graph connect the vertices/entities/nodes, and facts can be represented as an RDF[1] triple (*head, relation, tail*), which is also notated as <h,r,t>. As a result, a fact can be inferred by the relationship that connects two interrelated entities. Figure 3 demonstrates a sample KG demonstrating the semantic representation of entities captured from different interrelated healthcare domains, namely *Disease, Gene, Drug, and Compound*. The figure shows how a KG can be used to expand one domain by semantically interlinking it with another domain. Also, various facts can be inferred from the abstract structure of the KG. For example, the fact "*a Disease is associated with a Gene*" represents an abstract fact that comprises two abstract concepts (i.e. *Disease* and *Gene*), and the relation "*is associated with*" builds the triple <" *Disease*", "*associatedWith*"," *Gene*">. These abstract concepts can be then replaced with real-life entities to provide a specific domain representation. For example, the triple <" *Sjogren's Syndrome*",

[1] https://www.w3.org/TR/rdf11-concepts/

"*associatedWith*", "*HLA-DR3*"> indicates a fact about the Sjogren's Syndrome disorder which can be associated with HLA genes, namely HLA-DR3 [12].

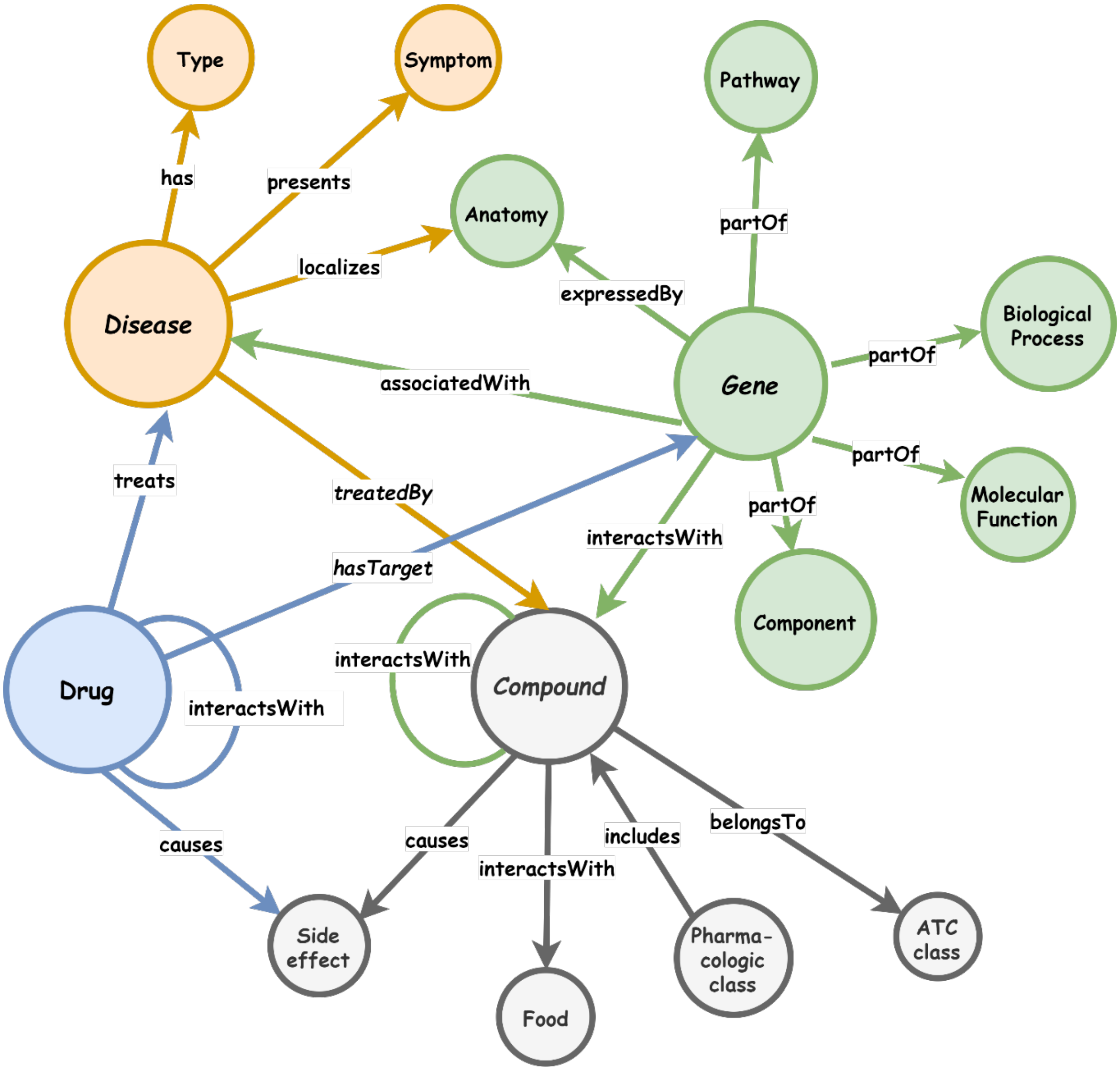

*Figure 3: A sample healthcare KG.*

The sample KG depicted in Figure 3 can be further expanded and linked with other datasets and vocabularies to extend the understanding of these real-world entities which belong to one or different domains.

## 3.2 Generic and Domain Specific KG

There are two types of KGs: generic and domain-specific KGs. Since the Semantic Web's inception, generic KGs (also called domain-independent, cross-domain, or open-world) have been constantly expanded. As a natural representation of interconnected entities, generic KGs have been related to linked data [13]. Cyc[2], BabelNet[3], NELL[4], CliGraph[5] , YAGO[6], and DBPedia[7] knowledge bases are examples of generic KGs, and the number of such KGs is rising rapidly. Domain-specific KGs are defined as "an explicit conceptualisation to a high-level subject-matter domain and its specific subdomains represented in terms of semantically interrelated entities and relations" [14]. These KGs are important to conceptualise specific domains, such as health, sports, social science, engineering, travel, etc. Examples of domain KGs include: HKGB [15], K12EduKG [16], SoftwareKG [17], ClaimsKG [18].

# 4. A taxonomy of healthcare KG construction

To better understand the overall paradigm of healthcare KG construction, we design a taxonomy that illustrates key activities and aspects of this process. Figure 4 shows the schematic representation of the taxonomy that was designed after careful examination of all significant state-of-the-art KG creation approaches relevant to critical healthcare applications, including (i) drug discovery, repurposing and adverse reaction; (ii) diseases and disorders; (iii) biomedicine; and (iv) other miscellaneous healthcare applications. This taxonomy aims to ensure that the process of constructing a typical KG in healthcare must demonstrate the intended primary use of KG, levels of knowledge extraction (entity level and relation level), different types of knowledge bases and sources, and evaluation metrics and criteria. The following sections provide detailed descriptions of each of the aforementioned elements.

## 4.1 Levels of knowledge extraction

The mechanism used to build a typical healthcare KG includes extracting entities and relations that can be captured from various heterogeneous healthcare data sources using a range of extraction methods. This section discusses the knowledge extraction procedures at both the entity level and the relation level.

### 4.1.1 Entity-level

Entities in healthcare KGs represent the nodes of the graph, which correspond to real-world entities such as drugs, diseases, diagnoses, patients, hospitals, events, etc. There are three main approaches used for entity extraction [19, 20]; (i) Named Entity Recognition (NER); (ii) Named Entity Disambiguation (NED); and (iii) Named Entity Linking (NEL). NER techniques aim to analyse textual data, thereby identifying factual names of various real-world objects. For example, the "*Pfizer*" entity in the following text snippet "*Clinical trials showed that Pfizer is effective.*" refers to the name of BioNTech vaccine that protects against COVID-19. The techniques used in NER can be classified into (a) knowledge-based techniques that rely on domain-specific knowledge and (b) advanced machine learning techniques that benefit from annotated data (in case of supervised learning), or partially annotated data (in case of semi-supervised learning), or derive knowledge from the structural or distributed nature of data (in case of unsupervised learning) to carry out an entity recognition task. Examples of ML-based techniques include Hidden Markov Models (HMM), Support Vector Machines (SVM), Conditional Random Fields (CRF) and variations, and Decision Trees [21-23].

[2] https://www.cyc.com/
[3] https://babelnet.org/
[4] http://rtw.ml.cmu.edu/rtw/kbbrowser/
[5] http://caligraph.org/ontology/Scientist
[6] http://www.foaf-project.org/
[7] https://wiki.dbpedia.org/

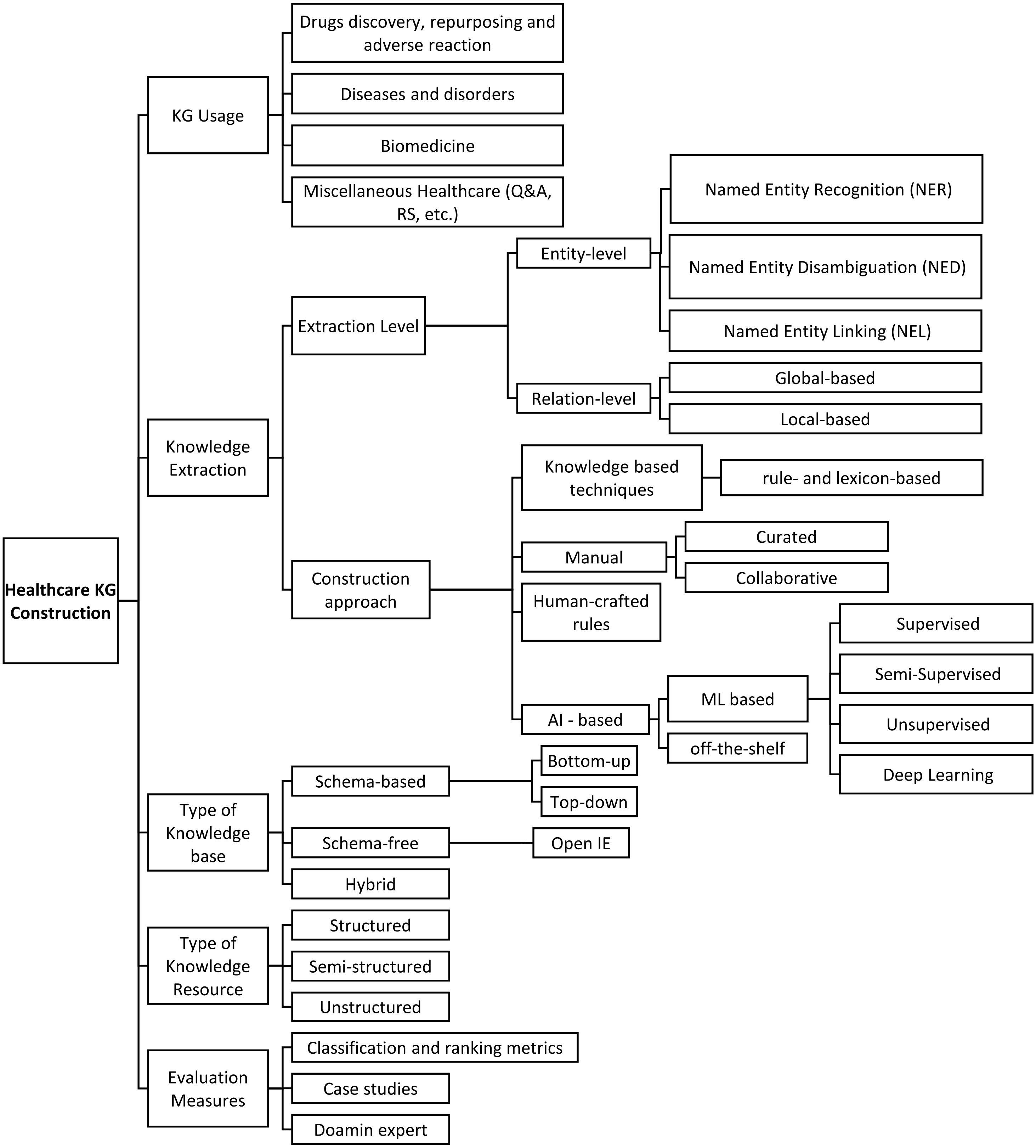

*Figure 4: A taxonomy of healthcare KG construction.*

Although NER techniques can identify potential entities, some of these units can be difficult to link to their corresponding entities that are located in the same or different KGs. For example, the expression "*Pink eye*" captured from any textual snippet could possibly refer to *conjunctivitis* and thus should be linked to a corresponding entity in a medical KG; or could simply refer to a cosmetic makeup term (*eyeshadow*) that relates to a completely different domain. The spectrum of currently used techniques in named entity disambiguation spans from rule-based approaches to advanced machine learning approaches, serving to clarify the results of NER and separate similar cases. Finally, NEL aims to link an identified entity (using a NER method) with an unambiguous manifestation (using a NED method) of the same entity captured from textual content, and frame it within a fixed context by linking it to a KG. As a result, NEL is the process of locating an entity mentioned in an (unstructured) text and linking it to a (structured) KG entry. The reader can refer to [19, 20] for detailed discussions on entity extraction mechanisms and technical issues related to their practical implementation.

#### 4.1.2 Relation-level

A relation between two entities conveys the semantic relationship between these entities. Extracting the relations between entities in KG requires such links to be identified, thus establishing a tuple that connects two potential entities. The aim of relation extraction is to figure out in which ways the identified and disambiguated entities are related semantically. This operation can be performed using either a local or a global strategy. The former denotes a mention-level relationship that is frequently inferred from short textual contents, while the latter seeks to infer relationships that span multiple knowledge bases and may involve numerous local relationships. Further information on relation extraction methods can be found in the related literature [24, 25].

### 4.2 Types of knowledge base

The course of construction of a healthcare KG is dependent on whether a predetermined ontology schema is used (schema-based), no predefined schema is used (schema-free) [26], or a combination of schema-based and schema-free techniques is employed. Based on the selection of data sources and ontology [27, 28], the first class of methods (schema-based) can be divided into two groups: (i) the bottom-up methods, in which the structural framework of an ontology is used as a foundation to construct the KG (e.g. Wikipedia is established by using the predefined ontology model, i.e. DBpedia [29]); and (ii) the top-down technique (e.g., YAGO) [30]), in which the ontology schema is inferred from the underlying structured data, or the taxonomies (hierarchy) which are developed based on information on the Web [31]. Schema-free methods are generally based on open information extraction strategies that rely on the open access to information on the Internet; as a result, data is gathered with diverse knowledge extraction techniques without particular concern for fitting the data into a unifying ontology design (e.g. OpenIE [32]). Hybrid knowledge-based approaches: are flexible strategies for obtaining knowledge that partially rely on a specified ontology but integrate new information in a flexible way (e.g. KnowledgeVault [27], NELL [33]).

### 4.3 Types of knowledge resources

Building a consolidated healthcare KGs requires extracting and integrating data from a variety of sources. The integration step is necessary in order to harmonise the data and provide a consistent big-picture view. There are three types of healthcare knowledge resources; (i) unstructured data sources (such as EMRs, medical literature, discharge summaries, and radiology reports); (ii) semi-structured or tree-structured data sources like JSONs and XMLs (e.g., Bio2RDF[8]); and (iii) structured databases that organize information in tabular formats such as relational medical databases (e.g., MEDLINE[9]).

### 4.4 KG evaluation metrics

The sudden growth of demand for healthcare KGs and the corresponding rush to produce them raises concerns about the quality of embedded information (i.e., entities and relations) and whether these elements accurately transmit the intended real-world facts behind the numbers. Assessing the completeness and veracity of information contained within a KG is the key to determining its "fitness of purpose" [34] for various downstream applications, as well as ascertaining data quality [35-38].

The lack of a complete and accurate KG in a particular domain makes the evaluation process difficult. This is due to the fact that compiling all factual data regarding a particular topic is a massive undertaking that may never be actually finished. As a result, several attempts have been made to augment and dynamically updateknowledge graphs with new facts derived from new entities and/or relations, usually referred to as KG Augmentation/Completion approaches. These efforts are subjected to correctness and completeness evaluation procedures to assure data quality. The evaluation can be performed by tracking classification accuracy and ranking metrics such as Hits@N and Mean Reciprocal Rank (MRR), Accuracy, Precision, Recall, and F-score [39, 40], based on a comparison between data in the KG and ground truth. These metrics are among a number of tools that can be used to assess the KG's construction quality and factuality of the described entities and relationships. Case studies and domain experts have also been occasionally used in the evaluation of KG structures [41, 42].

## 5. State of the art review

Recently, the Healthcare sector has gained much public attention, particularly with the coronavirus (COVID-19) pandemic that started in 2019 and continues to rattle the world. Therefore, there is a notable consensus between industry and academia that it is critical to consolidate the efforts of all stakeholders to overcome the challenges of this vital sector [43]. KGs offer the technical means to the healthcare sector to derive meaningful insights from voluminous and heterogeneous healthcare data contained in clinical and academic sources [44, 45]. The examined papers relevant to healthcare are classified into four different categories: (1) Drugs: This category comprises studies that incorporate KG technology for drug discovery, drug repurposing, and adverse drug reactions. (2) Diseases and disorders: Which includes studies that benefited from KG technology to conceptualise various diseases and conditions, such as stroke, subarachnoid haemorrhage, hepatitis, etc. Also, it includes papers about mental illnesses, such as depression, anxiety, autism, etc. (3) Biomedical studies: These include the fields of biomedicine, microbiology, etc. (4) Miscellaneous healthcare: These are works that span different categories, or those that incorporate KGs to model a specific healthcare solution.

### 5.1 Drug discovery, repurposing and adverse reaction

**Drug discovery:** KGs are receiving a lot of attention from researchers who are involved in the drug development studies. The necessity to construct specific KG for the drug sector has several key motivating factors [46]; prescribing a particular drug to treat a certain disease might involve some non-medical factors including the demographics, insurance policy, drug availability, etc. Further, in some instances, healthcare professionals who are not qualified to prescribe drugs, might act upon an emergency, thereby initiating a treatment that in a normal situation has to be initiated by a specialist doctor. Such complications illustrate the need for an intelligent platform that can actively guide the search for the optimal drug to prescribe. In this context, Mann et al. [46] attempted to create such a platform that can assist in finding a valid treatment considering the known symptoms or identified disease. In this study, the authors integrated existing medical knowledge resources, thereby building a KG to benefit the entire domain. In the same line of research, Che et al. [47] proposed a method to integrate six knowledge bases into one coherent KG. The resulting KG is then embedded into Graph Convolutional Network with an Attention mechanism for Drug–Disease Interaction (DDI), which is used to predict and discover potential drugs capable of effectively treating COVID-19. The prediction of Drug-Target Interaction and Drug-Drug Interaction are important aspects of the development of new drugs. In another interesting study, Zhang et al. [48] constructed two designated KGs describing drugs captured from a biological dataset, namely Bio2RDF[10]. This is followed by developing a learning model based on graph representation (MHRW2Vec), whose output was fed to a neural network model (TextCNN-BiLSTM Attention Network (TBAN)). The ultimate objective was to predict potential interactions of various COVID-19 drugs. Ye et al. [49] developed KGE_NFM, an integrated framework comprising both a KG and a recommender system to predict DDI. The components of the KG were embedded in a low-dimensional space, after which a neural factorization machine was tasked to build the recommender system for drug target discovery. Drug discovery incorporating KG technology was also discussed in [50-53].

---

[8] https://bio2rdf.org/

[9] https://www.nlm.nih.gov/medline/medline_overview.html

**Drug repurposing:** Drug repurposing (a.k.a. reprofiling, redirecting, rediscovery, or repositioning) is an interesting domain that has come into focus recently. It aims to reuse existing drugs to treat emerging diseases (such as COVID-19) thereby reducing both drug development timelines and the associated costs [54]. Therefore, various studies attempted to provide intelligent solutions for the challenges inherent in drug repurposing. Regarding the use of KGs, there is a direction of research aimed at constructing KGs that can be used for drug repurposing. BenevolentAI's proprietary KG [55] is amongst the most successful approaches in this research line. The BenevolentAI KG integrates an assortment of medical data obtained from structured and unstructured scientific repositories (including literature). It is queried by various algorithms to identify new relationships between entries, thereby suggesting new ways of treating diseases. In the same context, Wang et al., [56] proposed a framework called COVID-KG which aimed to construct a KG from multimodal data found in scientific literature into one actionable KG that can be used for drug repurposing. The proposed KG is built on an ontology of 77 entity subtypes and 58 event subtypes, as defined in the Comparative Toxicogenomic Database (CTD) (Davis et al., 2016), and entities are linked using Medical Subject Headings (MeSH) framework. [57] also proposed a multimodal drug repurposing KG for COVID-19 that was built with data harvested from scientific literature, and aimed to provide an overview of pathophysiology related to COVID-19. The construction of this graph was carried out manually using Biological Expression Language, and evaluated based on multiple case studies. Drug repurposing is further discussed in [58-63].

**Adverse drug reactions (ADRs):** ADRs refer to undesired reactions that occur after the use of a certain medical product [64]. ADRs carry significant risks to both patients and the hospital system [65], thus serious attention is required to tackle this issue and develop optimal technological solutions to mitigate it. The sophisticated structure of KGs presents an opportunity to define this problem conceptually and provides new ways to predict potential

[10] https://bio2rdf.org/

ADRs. Many studies were conducted in this direction, notably Bean et al. [66] benefited from access to two drug resources (namely DrugBank[11] and SIDER[12]) to build a KG that can predict ADRs. This KG contains four types of nodes and three types of edges, and is consolidated with a prediction model (similar to linear regression). Authors of [67] introduced a KG to represent drugs and ADRs, with data embedded using the Word2Vec model. On top of this model, logistic regression was used to predict whether a given drug causes any ADRs. Tumor-Biomarker Knowledge Graph (TBKM) [68] is another attempt to design a KG with four node classes (namely Tumor, Biomarker, Drug, and ADR) based on data from scientific biomedical literature. The aim of the KG is to discover ADRs of antitumor drugs as well as provide explanations why they occur. Predicting and discovering ADRs have been further reported in [69-74]. Zhao et al. [75] designed their drug action mechanism KG after extracting information from 770,000 abstracts of medical papers. Despite the poor approach used to extract entities and their relationships, the paper managed to cover a large number of drugs and mechanisms of action. Table 1 illustrates a summary of currently proposed KG construction approaches for drug discovery, drug repurposing, and adverse drug reaction.

*Table 1: A Summary of KG construction approaches for drug discovery, drug repurposing, and adverse drug reaction.*

| **Ref.** | **KG Specific Functionality** | **Knowledge Extraction Techniques** | | **Type of KB** | **KG Resource(s)** | **KG Stats** | **Evaluation Measure(s)** | **Shortcoming(s)** |
|---|---|---|---|---|---|---|---|---|
| | | **Entity-level** | **Relation-Level** | | | | | |
| [46] | Drug discovery | Manual and fuzzy matching | | Schema-based | Wikidata, DrugBank[13], WedMD, and GoodRx | N/A | R, P | • Lack of statistics on the resultant KG.<br>• Limited discussion on the Ontology design<br>• The evaluation of the proposed model emphasized on KG embedding rather than the resultant integrated KG. |
| [47] | Drug discovery for COVID-19 | Manual construction based on six KGs obtained from the literature | | Schema-based | Literature on COVID-19 | #n: 100,00<br>#e: 670,000 | AUC, and AUPRC | • Insufficient discussion on the mechanism followed to integrate the incorporated KGs,<br>• The evaluation of Att-GCN-DDI is limited and not detailed. |
| [48] | Drug discovery | Manual extraction based on Bio2RDF KG | | Hybrid | Bio2RDF[14] | #n: 2,947,140<br>#e: 10,131,654 | AUC, AUPR, F1 | • Inadequate discussion on the construction of drug KG. |
| [55] | Drug repurposing | Algorithms developed at BenevolentAI[15] and part of their IP | | Hybrid | Structured and unstructured resourced including Literature on COVID-19 | #n: millions<br>#e: hundreds of millions | Case study | • There is no detailed discussion on the mechanism followed to construct BenevolentAI graph.<br>• The evaluation was merely measured by case study. |
| [56] | Drug repurposing | Coarse- and fine-grained | Manually based on CTD and MeSH | Schema-based | Multimodal scientific literature (CTD[16]) | #n: 67,217<br>#e: 77,844,574 | Case study on Drug Repurposing | • Although the proposed framework demonstrated success in tackling the quantity issue of relevant KG resources, |

[11] https://go.drugbank.com/
[12] http://sideeffects.embl.de/
[13] https://go.drugbank.com/
[14] https://bio2rdf.org/
[15] https://www.benevolent.com/
[16] http://ctdbase.org/downloads/

| | | | | | | | | |
|---|---|---|---|---|---|---|---|---|
| | | entity extraction | | | | | Report Generation | the quality issue was not properly evaluated to demonstrate its effectiveness.<br>• Observed bias in training and development data, source, and test queries. |
| [57] | Drug repurposing | Manually encoded in Biological Expression Language | | Schema-free | PubMed, LitCovid[17], EuropePMC, etc. | #n: 4,016<br>#e: 10,232 | Case study (Gene Expression Analysis) | • The mechanism followed to construct the KG (manual-based) is poor in terms of scalability. |
| [58] | Drug repurposing | Cross-referencing | | Schema-based | PharmGKB, TTD, KEGG DRUG, DrugBank, SIDER[18], and DID | N/A | Case study (Finding drug–disease pairs) | • The proposed data model that was used for data integration can be improved by using formal domain ontology toward better conceptualizing the domain. |
| [67] | Prediction of adverse drug reactions | Direct construction from structural databases | | Schema-free | DrugBank database and SIDER database | #n: 12,473<br>#e:154,239 | P, R, F1, AUC, and a case study on Drug-induced liver injury | • The KG skips information of drugs and protein target,<br>• The scope of information perceived by entities can be enlarged by using longer path in the KG as the input of Word2Vec model. |
| [66] | Prediction of adverse drug reactions | Direct construction from structural databases | | Schema-free | DrugBank, SIDER | #n: 5,828<br>#e: 70,382 | AUC and case study(Validation in EHRs and Eudravigilance) | • No clear discussion on KG construction approach,<br>• Insufficient discussion on the methodology followed in the ML benchmark comparison. |
| [68] | Discovery of adverse drug reactions | cTAKES19 | naive Bayesian model | Schema-based | MEDLINE | #n: 9,699<br>#e: 139,254 | co-occurrence analysis and Case study (Osimertinib) | • The computed drug-biomarker groupings cannot differentiate between a drug-treatment relationship,<br>• The study lacks the attention to drug-drug interaction,<br>• lack of rationale on using the entity extraction method |
| [75] | Drug action | Automatically using rule-based approach | | Schema-free | Medical papers | #n: 40,963<br>#e: 57,865 | R, and accuracy | • Lack of verification to the textual prio KG construction.<br>• Limited comparison with currently exiting similar KGs. |

## 5.2 Diseases and disorders

**Topographic and anatomic:** KGs have accelerated the pace of scientific discovery that aims to better understand diseases affecting the human body. For example, Zhang et al. [15] developed Health Knowledge Graph Builder (HKGB), which is a framework that can be used to construct a Health KG (HuadingKG) from multiple sources (namely EMRs, medical standards, and expert knowledge) to be used in the cardiovascular domain. To conceptualise subarachnoid haemorrhage stroke, the authors of [45] developed a comprehensive framework that allowed them to construct a KG from heterogeneous data automatically. In particular, the authors incorporated semantic analysis for entity and relation extraction, and implemented a knowledge prediction model based on the association rule and ensemble machine learning. KGHC [76] is a KG designed specifically for Hepatocellular Carcinoma. It brings together and connects entities captured from 5 different unstructured and structured data sources and extracted using information extraction techniques such as BioIE and SemRep. Yin et al. [77] constructed a KG for diagnosing and treating viral hepatitis B by adopting a top-down approach where a domain ontology was used to build the KG. The authors did not provide adequate details on the mechanism utilised to construct the KG or the evaluation metrics. Yet, they claimed that the designed KG benefits intelligent recommender systems that can be used to diagnose and treat viral hepatitis B. Another research direction identified the role of genes in human disease [78]. The authors built a convolutional neural network-based model on top of a biological KG to classify the genes highly correlated with cancer. While the construction of the KG itself was not adequately validated, the resultant embedding model was evaluated on downstream tasks. An attempt at preventing Myopia using KG technology was described in [79]. The authors developed a KG from various Chinese websites to provide intelligent Q&A services to users interested in Myopia prevention. However, the finalised KG lacks multimodal data that can be captured from medical databases and domain-relevant question answering systems. Conceptualising Stroke and its causes and effects is an extensively covered subject in the literature. For example, Yang et al. [80] constructed an integrated KG, named StrokeKG, that portrays various stroke-relevant relationships inferred from various medical datasets. Designing KGs to benefit the victims of stroke was also examined in [81, 82]. COVID19-related disease discovery using KGs was reported by Huang et al. [83]. Relying on a pipeline approach, the authors surveyed from relevant scientific papers related to COVID-19 and used the collected data to construct a KG that can identify diseases and drugs associated with COVID-19. The accuracy of the extracted knowledge was then verified using the time-slicing method. The use of KGs in the healthcare domain was discussed with a focus on disease identification and prediction in [84-87], detecting the association between miRNA and disease in [88], chronic disease management in [89], and syndromes diagnosis in [90].

[17] https://www.ncbi.nlm.nih.gov/research/coronavirus/
[18] http://sideeffects.embl.de/
[19] https://ctakes.apache.org/

**Mental disorders:** Yuan et al. [91] constructed a KG with minimal supervision to frame autism spectrum disorder diseases, using the articles obtained from the PubMed dataset. Entities were extracted using MinHash lookup/ UMLS [92] and formed into pairs which were then clustered using kmeans++ based on similarity between entities. Constructing KGs that can describe depression was undertaken by Huang et al. [93]. In particular, they attempted to generate a sub-graph that describes depression disorder, obtained by parsing data from a variety of major knowledge sources such as PubMed, Medical Guidelines, DrugBank, Unified Medical Language System (UMLS) etc. In the same line if research, Li et al. [94] proposed a UMLS-based semantic prediction program, known as SemRep, as well as SemMedDB to construct a KG for describing depression by using a bottom-up approach. Depression and its association with metabolism is also discussed in [95]. The authors developed MDepressionKG KG that integrates data about human microbial metabolism network, human diseases, microbes, etc., to offer semantic-based rational reasoning and establishing probable relations between depression and comorbid diseases. Although the authors furnish a useful online website to demonstrate utility of MDepressionKG, the knowledge inference mechanism is ineffective due to the incorporated traditional rules of logic. Furthermore, automatic extraction methods are required to enrich the functional diversity of the proposed depression KG. The conceptualisation of various mental disorders through graphs was also presented in [96-99]. Table 2 shows a summary of KG construction approaches for diseases and disorders.

*Table 2: A Summary of KG construction approaches for diseases and disorders.*

| Ref. | KG Specific Functionality | Knowledge Extraction Techniques | | Type of KB | KG Resource(s) | KG Stats | Evaluation Measure(s) | Shortcoming(s) |
|---|---|---|---|---|---|---|---|---|
| | | **Entity-level** | **Relation-Level** | | | | | |
| [15] | Cardiovascular domain | LSTM-CR | pattern-based and supervised learning methods | Hybrid | UMLS, EMRs, medical standards, and expert knowledge. | #n: 8,293,284 #e: 32,256,360 | The evaluation is conducted in the embedded modules | • The overall framework requires a detailed case study to evaluate the effectiveness of integrating the proposed modules. |
| [45] | Subarachnoid hemorrhage | Semantic analysis (Ontologies: LBO, IAO, etc.,) | Automatic (Rule-based) | Shema-based | clinical notes and brain angiograms | N/A | P, R, F1, and AC | • Limited discussion on the KG statistics<br>• The overall framework requires a detailed case study to evaluate the effectiveness of integrating the proposed modules. |
| [76] | Hepatocellular carcinoma | SemRep[20], rule-based method,and BioIE(with Att-BiLSTM-CRF) | | Schema-based | PubMed, SemMedDB, UpToDate, and Clinical Trials[21] | #n: 5,028 #e: 13,296 | Accuracy | • The KG was not properly evaluated on real-life case study that addresses hepatocellular carcinoma.<br>• There has been no detailed discussion on the mechanism followed to address the presented disagreements. |
| [80] | Stroke | DNorm[22], tmChem[23], GNormPlus[24], PWTEES[25] | NLTK, PKDE4J, and Bio-BERT | Shema-free | CID[26], TCMID[27], EU-ADR[28], ETCM[29] | #n: 46 k #e: 157 k | P, R, F1 | • The constructed KG is limited to Chinese context and hard to replicate and build a more comprehensive map of medical knowledge. |
| [77] | Diagnosis and treatment of viral hepatitis B | N/A | N/A | Schema-based | EMR (8544 patients in China) | #n: 8,563 #e: 96,896 | N/A | • No proper evaluation was conducted.<br>• No discussion on mechanism followed to construct the KG |
| [83] | Coronavirus pneumonia-related diseases, | CRF | Bio-BERT | Shema-free | COVID-19 scientific literatures | #n: 10,993 #e: 1,204,234 | Specificity, P, R, F1, and AC | • The entity and relation extraction datasets are provided with lack of discussion on the mechanism followed to conduct the experiments on these datasets. |

[20] https://semrep.nlm.nih.gov/
[21] https://clinicaltrials.gov/
[22] https://www.ncbi.nlm.nih.gov/CBBresearch/Lu/Demo/DNorm/
[23] https://www.ncbi.nlm.nih.gov/research/bionlp/Tools/tmchem/
[24] https://www.ncbi.nlm.nih.gov/research/bionlp/Tools/gnormplus/
[25] https://github.com/chengkun-wu/PWTEES
[26] http://www.cbs.dtu.dk/services/
[27] http://bidd.group/TCMID/
[28] https://biosemantics.erasmusmc.nl/index.php/resources/euadr-corpus
[29] http://www.tcmip.cn/ETCM/

| | | | | | | | | |
|---|---|---|---|---|---|---|---|---|
| [78] | Identifying disease-gene associations | N/A | N/A | Shema-free | CTD, BioGrid30, MalaCards31 | #n: 103,625<br>#e: 3,273,215 | N/A | • No discussion on the mechanism followed to extract entities and relationships.<br>• The construction of the KG itself is not evaluated |
| [79] | Myopia Prevention | Automatic using python script | | Schema-based | Baidu Encyclopedia, Chinese Wikipedia, and professional websites | #n: N/A<br>#e: N/A | NA | • KG is not described in terms of mechanisms used to extract entities and relationships.<br>• No proper evaluation is undertaken. |
| [93] | Depression disorder | XMedlan, Semantic Queries with regular expressions, | | Hybrid | PubMed, Clinical Trials[5] DrugBank[32], DrugBook, Wikipedia, SIDER[33], and UMLS | #e: 8,892,722 | Use cases | • Lack of proper evaluation,<br>• insufficient use of other important medical repositories,<br>• lack of discussion on both the methodology used for knowledge integration and KG statistics. |
| [91] | Autism spectrum disorder | MinHash lookup/UMLS | Skip-gram and kmeans++ | Schema-free | PubMed[34] (autism spectrum disorder-related article abstracts) | #n: 6827<br>#e: 16,192 | Hit@k | • Extracted relations are coarse-grained.<br>• Difficult to distinguish semantically related relations,<br>• Insufficient overall evaluation to the model |
| [94] | Depression | SemRep[35],OpenIE and rule-based method | | Schema-based | SemMedDB, PubMed | #n: 3,055<br>#e: 30 | Jaccard | • Poor data quality<br>• The utility of KG was not well-proven |
| [95] | Metabolism-depression associations | Manual curation and extraction by domain expert (traditional logical rules) | | Schema-based | KEGG and scientific literature | #n: 3,724,526<br>#e: 5,725,821 | Case study | • Ineffective inferences due to the incorporated traditional logical rules.<br>• Automatic extraction methods are required to enrich the functional diversity of the depression KG. |

## 5.3 Biomedicine

**Generic biomedicine:** KG's have been used with success for modelling both biological systems and pathologies, providing the means to understand this interplay between them. Several studies reported significant advances in this direction while incorporating KG technology. PharmKG [100] is a comprehensive KG built upon integrating six interrelated knowledge bases, with nodes representing genes, chemical compounds, and diseases. Entities of PharmKG are labeled with domain-specific information, keeping semantic and biomedical characteristics of the data. Percha et al. [101] compiled a basic KG known as the Global Network of Biomedical Relationships (GNBR) from biomedical literature. The process of populating GNBR with data was performed using PubTato (named entity annotator) tool, as well as Ensemble Biclustering for Classification (EBC) algorithm to annotate entities captured from Medline abstract. Wood et al. [102] developed RTX-KG2, an integrated KG that includes biomedical data captured from 70 biomedical knowledge bases. The aim of RTX-KG2 is to offer an open-source KG that can be used as a biomedical translational reasoning engine. Zhang et al. [103] reported the extraction of biomedical causality from the scientific literature. In their work, the authors constructed a biomedical knowledge graph to discover causal relationships in the biomedicine field. Authors of [82] developed a marine Chinese medicine KG using a top-down approach that takes guidance from a domain ontology. Developing an integrated KG to benefit the biomedical domain has also been discussed in [104]. The authors presented BioKG, a KG of drug-drug and drug-protein interactions data collected and compiled using modular software, namely BioDBLinker. This KG contains managed entities and relationships captured from at least five biomedical databases, such as UniProt, REACTOME, KEGG, DrugBank, SIDER, and Human Protein Atlas (HPA). He et al. [105] designed a KG for intestinal cells. First, the authors built an ontology as a conceptual model followed by extracting facts from the academic literature. Despite the problems with mechanisms used to construct the actual KG, the work presents an important attempt toward constructing KGs specifically to study the intestinal field, facilitating much easier observation of the processes of intestinal cytokines via various signalling channels. Constructing KGs to benefit generic biomedical domain was the subject of [106-111].

[30] https://downloads.thebiogrid.org/BioGRID
[31] https://malacards.org/
[32] https://www.drugbank.ca/
[33] http://sideeffects.embl.de/
[34] https://pubmed.ncbi.nlm.nih.gov/
[35] https://lhncbc.nlm.nih.gov/ii/tools/SemRep_SemMedDB_SKR/SemRep.html

**Microbiology:** KGs offer an excellent mechanism to conceptualise our understanding of microscopic organisms and their ecological traits. To this end, Joachimiak et al. [112] developed KG-Microbe, an integrated KG that contains prokaryotic data for phenotypic traits as well as supporting use cases in microbiology, biomedicine, and environmental science. Liu et al. [113] conceptualised gut microbiota using a semantically enriched KG, namely MiKG. MiKG integrates facts obtained from medical literature as well as other medical knowledge bases, thereby offering an interface for detection of possible connections between gut microbiota, neurotransmitters, and mental disorders. Authors of [114] developed a Microbe-Disease Knowledge Graph (MDKG) through an explorative study, thus identifying the associations between bacteria and diseases. MDKG is populated with entities and relations captured from textual content of Wikipedia as well as other semantic knowledge bases. Modelling Coronavirus using KG technology has recently attracted a lot of attention in the research community. For example, Zhang et al. [115] built a coronavirus KG by integrating entities captured from Analytical Graph and CORD-19 databases. The aim of the proposed KG is to provide a tool for the exploration of coronavirus on the entity level. Another attempt to help the biomedical research community comprehend the coronavirus using KGs is offered by Chen et al. [116]. The authors constructed a designated KG to discover any associated diseases, potentially effective drugs or treatments, and relevant genes and mutations. Using KG technology, modelling Coronavirus relevant information was also implemented and discussed in [57, 117, 118]. Further uses of KGs in microbiology are studied in [119]. Table 3 depicts a summary of KG construction approaches for the biomedical domain.

*Table 3: A summary of KG construction approaches for the biomedical domain.*

| Ref. | KG Specific Functionality | Knowledge Extraction Techniques | | Type of KB | KG Resource(s) | KG Stats | Evaluation Measure(s) | Shortcoming(s) |
|---|---|---|---|---|---|---|---|---|
| | | Entity-level | Relation-Level | | | | | |
| [100] | Generic biomedicine | Manual integration and mapping of entities and relationships | | Schema-base | OMIM, DrugBank, PharmGKB, Therapeutic Target Database], SIDER, and HumanNet | #n: 7,603<br>#e: 500,958 | Hits@N and Downstream tasks | • The quality and integrity of the metadata cannot be fully assured.<br>• The final version of the constructed graph does not have large-scale of entities compared with state-of-the-art KGs. |
| | | | | | | | | • No discussion is provided on the adopted ontology. |
| [101] | Generic biomedicine | PubTator[36] and manual annotation (EBC) | Stanford Dependency Parser[37] | Schema-free | Biomedical literature (Medline abstracts[38]) | #n: N/A<br>#e: 2,236,307 | Benchmark comparison | • Heavily dependent on the co-occurrence of paths to map scarcer paths to themes,<br>• Lack of handling complex relations<br>• There is a potential of a parser error, |
| [102] | Translational biomedicine | Manually and automatically using Snakemake[39] | | Schema-base | 70 knowledge sources including SemMedDB, ChEMBL, etc. | #n: 6.4 m<br>#e: 39.3 m | Benchmark comparison | • The automation process to construct the KG was not detailed.<br>• The comparison with other KGs is not well discussed nor formulated. |
| [103] | Biomedical Causal Discovery | Manual and rule-based approach | | Schema-free | PubMed | #n: N/A<br>#e: N/A | Accuracy | The paper failed to extract implicit causality,<br>The process to identify concepts and relationships between concepts is not detailed. |
| [82] | Marine Chinese medicine | Manual mapping between the ontology and the KG | | Schema-base | Medical literature | #n: N/A<br>#e: N/A | NA | • The paper inadequately described the construction and evaluation of the proposed KG. |
| [104] | Generic biomedicine | BioDBLinker | Automatic mapping | Schema-free | UniProt[40], REACTOME[41], KEGG[42],DrugBank, SIDER, and d Human Protein Atlas (HPA)[43]. | #n: N/A<br>#e: N/A | Benchmark comparison | • Suffers from sparsity of data,<br>• Train-test data leakage in case used without careful review |
| [105] | Intestinal cells | Manually based on the conceptual model | | Schema-base | PubMed | #n: 2443<br>#e: 160253 | Case study | • Poor entity and relation extraction approaches.<br>• Data source is static and limited to medical literature, yet medical facts of intestinal cells can be obtained from future experiments. |
| [112] | Microbiology | NER and NLP techniques | | Schema-base | KG Hub – COVID19[44] | #n: 266,000<br>#e: 432,000 | N/A | • Poor discussion on mechanisms followed to construct and validate the KG |
| [113] | Gut microbiota | Manual annotation and mapping | | Schema-base | Google Scholar and PubMed, UMLS, MeSH, SNOMED CT, and KEGG | #f: 31,268,998 | Case studies | • Poor extraction of entities and relations.<br>• The correctness and completeness of extracted relations limit the semantic search's precision and reliability. |

36 https://www.ncbi.nlm.nih.gov/research/pubtator/
37 https://nlp.stanford.edu/software/lex-parser.shtml
38 https://www.nlm.nih.gov/bsd/pmresources.html
39 https://snakemake.readthedocs.io/en/stable/
40 https://www.uniprot.org/
41 https://reactome.org/
42 https://www.genome.jp/kegg/
43 https://www.proteinatlas.org/
44 https://github.com/Knowledge-Graph-Hub/kg-covid-19

| | | | | | | | |
|---|---|---|---|---|---|---|---|
| [114] | Microbe-Disease Associations | Kindred entity and relation classifier[45] | Schema-free | Wikidata, UMLS, NCBI | #n: 9,832<br>#e: 21,905 | Hits@N | • KG can be expanded by means of a bacterial attribute mining tool,<br>• Lacks a discussion on interactions between bacteria and antibiotics or viruses. |
| [115] | Coronavirus | Manual extraction and mapping | Schema-free | Analytical Graph (AG) and CORD-19[46] | #n: 588,820<br>#e: N/A | Case study | • Limited data sources,<br>• Static KG |
| [116] | Coronavirus | BioBERT | Schema-free | PubMed and CORD-19 | #n: N/A<br>#e: N/A | P, R, and F1-score | • KG can be expanded to other bio-medical datasets.<br>• Further biomedical NLP models for NER, e.g., blueBERT can be attempted to verify the validy of the extracted knowledge. |

## 5.4 Miscellaneous Healthcare

**Constructing KGs from EMRs:** The ongoing efforts to leverage the proliferation of EMRs for multiple medical applications are well-documented in the scientific literature. Extracting valuable knowledge from such data silos has been made easier by KG technology. In this context, several studies attempted to construct medical KGs that can improve specific areas, for example, clinical decision support systems. One such attempt was undertaken by Li et al. [120], who followed a systematic approach consisting of eight steps to build a medical KG from EMRs obtained during the patients' visits. The authors constructed a quadruplet-based medical KG incorporating an additional item (properties) which includes a set of characteristics to rank the embedded entities. The main objective of this study is to ensure the robustness of facts in the KG related to the medical domain. Evaluating the robustness of a constructed KG in healthcare is of utmost significance to ensure the quality of the inferred knowledge. In this context, [121] presented a methodology to measure and evaluate the robustness of knowledge relating to diseases and symptoms, with data captured from existing health knowledge graphs as well as records of patient visits to the Beth Israel Deaconess Medical Center (BIDMC). Postiglione et al. [122] proposed an advanced entity recognition approach named PETER (Pattern-Exploiting Training for Named Entity Recognition), that integrates Pattern-Exploiting Training (PET) [123] to build an Italian-language KG for healthcare. EMRs represent a fertile source of information for healthcare KGs, hence their use for construction of KGs is becoming quite common, as exemplified in [124] and [125].

**Query Answering (QA) and Question and Answer (Q&A):** Incorporating health KGs into a QA system was discussed by Sahu et al. [126]. The authors proposed a system that can be used to search for various health-based KGs and obtain a set of healthcare-related response sub-graphs. The possibility of using medical KG's to benefit QA applications was also discussed in [127]. Zhao et al. [128] made use of EMRs obtained from hospital patient records in Shanghai to build a medical knowledge graph based on the BILSTMCRF model. Here, a KG is used as a part of a QA system to provide support for establishing medical diagnosis. Xie et al. [129] attempted to create a KG for Traditional Chinese Medicine (TCM), yet the KG they ended up with is very limited in terms of entities and relationship; thus, the applicability and utility of the graph is questionable. Another Chinese medical KG was proposed by [130]. The authors developed this KG from various structured, semi-structured, and unstructured resources and built a QA system that was not adequately validated due to irrelevant results. Also, Huang et al. [131]

[45] https://kindred.stanford.edu/

[46] https://www.kaggle.com/datasets/allen-institute-for-ai/CORD-19-research-challenge

designed a QA system based on a constructed KG, with the information in the graph used to identify the question's intention. Deploying QA and Q&A systems based on medical and healthcare KGs is also a relevant topic in [132].

**Healthcare Management**: In the literature, it has been frequently suggested that a KG can be built to help with health management and to better address the most critical health-related issues and chronic disorders [133-136]. For example, Huang et al. [133], proposed a KG building approach that aids users who are seeking information about a healthy diet. Domain ontology was presented by the authors as the basic structure of a KG containing information about diet. Conditional Random Fields (CRF), Support Vector Machine (SVM), and Decision Tree (DT) methods were used to enrich the KG with entities harvested from a variety of healthcare websites. Haussmann et al. [134] developed an integrated KG (FoodKG) that brings together information about healthy food, recipes, and nutritional value. The authors used the RDF Nano publication to establish the reliability of their findings [137]. Chi et al. [135] developed an inclusive healthy diet KG by following a similar study path. In this case, the KG was comprised of five essential concepts: the meal, the dish, the nutritional aspect, the symptom, and the crowd. The proposed model was able to collect and import entities from a range of web resources and deployed multiple NLP and machine learning methods with a semi-automated extraction strategy. In addition, food domain-specific KGs were modelled in [138-140]. Another example of the use of KG-based technology to address difficulties in healthcare systems was discussed in [141-143].

**Miscellaneous KGs:** In healthcare, addressing the timing factor in KG creation is critical. Ma et al. [144] developed a temporal KG that is useful for studying episodic memory in cognitive tasks. The Integrated Conflict Early Warning System (ICEWS) dataset and the Global Database of Events, Language, and Tone were used to create this temporal KG (GDELT). Their work was unique in that it involved four substantial static KGs embedding data to four-dimensional temporal/episodic KGs, which set them apart from other efforts in this direction. Two new RESCAL generalisations were also proposed and considered. Another important effort that integrated plausible reasoning with fine-grained biomedical ontologies to tackle the data incompleteness problem was undertaken by Mohammadhassanzadeh et al. [42]. The authors proposed a Semantics-based Data analytics (SeDan) framework that performs an exploratory analysis of the KG using the OWL extension and query rewriting algorithm. The framework incorporates data from various knowledge bases, including the DrugBank, Disease Ontology, and the large-scale semantic MEDLINE database (SemMedDB). Rastogi et al. [145] framed their personal health KG as a combination of context, personalization, and integration with other knowledge bases. Their study indicated that the literature on personalised health-related KGs is incomplete and lacks a unified standard representation to adequately describe the designated domain. To provide an overview of effective medications, side effects, and target populations relevant to COVID-19, the authors of [146] proposed a KG-based framework to support COVID-19 clinical research. This framework benefited from Stanford's Stanza toolbox to extract KG's entities and relationships that can be fed into a visualisation module for querying information. The application of KGs in healthcare and medical domains was documented in other relevant tasks including epidemic contact tracing [147], food waste detection [109], drug similarity [148], clinical decision support systems [120], and medical recommender systems [149, 150]. Table 4 shows a summary of KG construction approaches used in various miscellaneous healthcare applications.

*Table 4: A summary of KG construction approaches for miscellaneous healthcare*

| Ref. | KG Specific Functionality | Knowledge Extraction Techniques | | Type of KB | KG Resource(s) | KG Stats | Evaluation Measure(s) | Shortcoming(s) |
|---|---|---|---|---|---|---|---|---|
| | | Entity-level | Relation-Level | | | | | |
| [120] | A generic medical KG of patient visits. | BMM, BiLSTM-CRF and pattern recognizer | Nine predefined relations | Schema-free | Southwest Hospital in China: 16,217,270 de-identified visits of 3,767,198 patients | #n: 22,508<br>#e: 579,094 | R, P, F1, and NDCG | • KG embedding was designed and limited to Bi-LTSM without considering other state-of-the-art techniques.<br>• The evaluation was mainly conducted on the embedded components.<br>• Besides the preliminary discussion on the applications, there is a lack of an overall evaluation of the KG. |
| [121] | KG of online EMR and emergency department | N/A | N/A | Schema-free | BIDMC dataset and EMRs from an emergency department | #n: N/A<br>#e: N/A | F1 and the area under the precision-recall curve | • The provided statistics are on the sources of the KG; the stats on the KG in terms of entities and edges are missing.<br>• There is no discussion on the mechanism followed to construct the KG in terms of entities and relations. |
| [133] | Smart Healthcare Management | CRF | Manual and classification-based algorithms | Schema-based | Chinese healthcare websites[47][48,49] | #n: 1,169<br>#e: 9,707 | R, P, and F1 | • The resultant KG can be consolidated with information about disease and drugs and link them with symptom entities. |
| [128] | Q&A | BILSTM-CRF | Manually | Schema-free | EMRs from a hospital in Shanghai | #n: 44,111<br>#e: 203,308 | R, F1 and Accuracy | • Lack of comparative study of the model.<br>• Limited practicability of the system<br>• Limited size and pretreatment of the corpus |
| [129] | Q&A | BiLSTM + CRF | | Schema-free | National Service Platform for Famous Old Chinese Medicine Experience[50] | #n: N/A<br>#e: N/A | Case study and Hitration | • Poor KG with a minimal number of entities and relationships, |
| [42] | Q&A | Plausible reasoning | | Schema-free | BioASQ, DrugBank, Disease Ontology, and SemMedDB | #n: N/A<br>#e: N/A | Domain expert's verification | • Insufficient evaluation,<br>• evaluating the performance of query rewriting algorithm does not exist |
| [130] | Q&A | Automatic mapping | | Schema-free | Chinese medical websites | #n: 18,687<br>#e: 88,858 | Case study | • Poor discussion on extraction of entities and relationships.<br>• The QA system does not exhibit utility due to inapplicable results. |
| [131] | Q&A | Jieba51 | Automatic mapping | Schema-free | A medical company (YiFeng Pharmacy[52]) | #n: 34,788<br>#e: 601,475 | Training and decision | • The construction of KG is not validated. |

47 https://www.zhys.com/
48 http://app.huofar.com/grz/
49 http://www.cf555.com/
50 https://www.gjmlzy.com/
51 https://pypi.org/project/jieba/
52 http://www.yfdyf.cn/

| | | | | | | | | |
|---|---|---|---|---|---|---|---|---|
| | | | | | | | accuracy, cost, and time | • The system can answer one intention per question and cannot thus answer questions with multi-intensions. |
| [146] | COVID-19 Clinical Research | Stanza's NER[53] | Stanza's Bi-LSTM | Schema-free | Artificial Intelligence in Medicine | #n: N/A<br>#e: N/A | Baseline comparison | • Lack of statistics on entities and relationships,<br>• Poor KG validation method |

## 5.5 Summary

This paper examines the most recent works related to KG construction methodologies in various healthcare domains, such as drugs (and their applications), diseases and disorders, biomedicine, etc. A closer look into these important domains reveals crucial research areas that benefit substantially from KG technology. This research demonstrates the popularity of using KGs to solve real-world healthcare-related problems and shows how KGs have proven to be an effective overall solution for reducing complexity, ensuring flexibility, and establishing a common-ground architecture where data from various sources can be readily incorporated. It is generally agreed that KG technology allows for semantic integration of data acquired from many sources, which may exist in different formats, and can then be fed into a single, coherent framework to be formally used to conceptualise the designated domain.

# 6. Findings, open issues, and opportunities

Despite the popularity of KG technology in the healthcare domain, this study reveals certain limitations that open new directions for future research.

- **KG data sources:** various previous studies have concentrated on knowledge curation and facts captured from a limited number of data sources. For example, certain KGs were constructed using only biomedical scientific publications (e.g. PubMed and SemMedDB) [94, 103, 105]. The extracted knowledge using such data sources lacks completeness, leading to poor descriptiveness of the entities and potentially flawed relationships within a particular healthcare domain. This also limits the capacity of the graph to deliver useful facts or rules to power data-driven methods that can be used for making healthcare decisions [45]. To consolidate a healthcare KG and establish a cohesive viewpoint of the domain, alternative sources need to be incorporated and integrated including EMRs, PMRs, clinical trials, patient records, epidemiological surveillance, sensor data, disease registries, wearable devices, health workforce data, census data, implanted equipment, pill cameras, and all other relevant sources. However, full integration of such heterogeneous data sources can be a complicated and time-consuming task, especially when working with large-scale datasets where traditional data assimilation and aggregation techniques are not applicable. Therefore, there is still room for research to address the big data problem in healthcare KGs by developing advanced and sophisticated data collection and aggregation techniques.

- **Healthcare knowledge interoperability:** Linked Open Data (LOD) and Semantic Web technologies have made it possible to improve a variety of domain-specific applications [14, 151-153]. KGs represent an expansion of these efforts and are frequently connected with LOD initiatives because they improve data semantics by enhancing the conceptual representations of entities [154]. As a result, appropriate interlinking of entities gathered from different data sources facilitates information interoperability, resulting in multimodal KGs. However, some of the methodologies investigated in this study revealed difficulties in attaining the appropriate level of knowledge expandability and interoperability. In particular, semantic expansion strategies were underutilised, and their ability to take advantage of freely accessible vocabulary and semantic resources is mostly ignored. The expansion of healthcare knowledge with health records collected from different channels, such as hospital admissions, family physician visits, prescription drugs, pharmacy requests, laboratory blood analyses, and death certificates establishes a comprehensive individual health (or disease) profile [155]. This holistic view carries enormous implications for several research areas, such as epidemiology and precision medicine. Basic structure of KGs facilitates better data integration, unification, and information sharing. Semantic expansion adds context to the collected facts in the KGs and enhances the quality of the aggregated knowledge, eliminates redundant records, and detects missing entities. Based on success of existing healthcare semantic expansion initiatives such as the Centre for Health Record Linkage (CHeReL) in Australia [156] and Rochester Epidemiology Project in USA [157], more research in this direction should be conducted.

[53] https://github.com/stanfordnlp/stanza

- **KG construction mechanisms:** The construction of the KG comprises several activities which might vary depending on the type of knowledge base (schema-based, schema-free, or hybrid), knowledge resources and their data types (structured or unstructured), knowledge extraction techniques (entity-level and relation-level), etc. Several of the examined studies failed to adequately disclose the internal mechanisms they used to build and implement the KGs. A shortcoming that was commonly observed was poor and/or limited discussion to explain either the overall construction methodology [48, 55, 66] or the essential construction tasks such as the ontology design [46, 100], entity and/or relation extraction [78, 83], and knowledge integration [47, 93]. Furthermore, many of the KGs described in those papers are not publicly available for inspection. These drawbacks detract from knowledge sharing, translation, and reusing, and make the replication of the proposed approaches difficult. This is particularly problematic in the healthcare domain where knowledge replicability can assist in consolidating the facts about certain scientific tests and medical experiments [158]. Therefore, future studies must ensure that all steps of KG construction are well-explained, and the resultant KG must be publically shared with the community to reinforce FAIR principles (Findable, Accessible, Interoperable, Reusable)[54].

- **KG evaluation:** Despite the continuous propagation of KGs for the healthcare domain and its sub-domains, this survey reports evident problems with KG evaluation and/or case study implementation. Numerous KGs were constructed with no proper concern for evaluation of their quality [77-79, 82]. Additionally, there is only a limited utility in applying the constructed KGs to real-life applications. Instead of practical applications, the proposed KGs mainly attempted to provide an underlying conceptual structure of the domain utilising domain-specific entities, concepts, relationships, and events. For example, the authors of [76] attempted to build a KG for hepatocellular carcinoma with no verified utility in addressing the designated disease. Designing and implementing actionable healthcare analytics must be the essence of the

[54] https://www.go-fair.org/fair-principles/

KG construction philosophy, where relevant facts are obtained with the objective to conceptualise the correct context and address a domain problem, thereby achieving the hoped-for value. Future works must ensure that KGs are assessed using one or more appropriate evaluation and refinement methodologies such as (i) silver and gold standards [159]; (ii) theoretically proven computational measures such as precision and recall; and (iii) domain experts. In addition, the constructed KG must prove its utility and verify its applicability in real-life scenarios and for the execution of downstream tasks.

- **Data Quality and Privacy**Applying healthcare KGs to downstream tasks such as drug discovery, clinical decision support, and medical treatment relies profoundly on the high quality of the embedded facts. Although some of the examined works constructed their KGs using structural, verified and curated data sources [42, 94, 104], other KGs imported data from unstructured sources (such as scientific medical literature or social media), with little regard for applying data quality measures before incorporating the extracted information [56, 105]. Freely available texts such as scientific medical literature commonly comprise ambiguous data, abbreviations, and noisy data that includes words and phrases irrelevant to the designated context. EMRs also comprise a vital source of embedded clinical data that can be either mistakenly neglected or hard to collect due to confidentiality constraints. These challenges raise concerns about the quality and reliability of KGs generated from such data sources. Therefore, high-quality healthcare KGs should be constructed by selecting high-quality data sources and developing quality measurement techniques. Also, advanced NLP and deep learning algorithms that can efficiently and automatically identify high-quality entities and relations should be implemented wherever possible. Those tools should be used to improve data privacy, integrity, and security, preventing malicious activities that attempt to abuse patients' sensitive medical information.

- **Recentness**: Most of the examined studies did not consider the temporal factor; their KGs are static in nature and often neglect the validity period of incorporated triples. A healthcare KG built based on just one snapshot of the knowledge landscape might not be a sustainable depiction of the designated domain, particularly with the emergence of wearable medical devices, sensors, health monitoring systems, and mobile applications [160] which make the construction of dynamic and frequently updated KGs a necessity. Ignoring the dynamic nature of healthcare knowledge degrades the quality and accuracy of facts embedded in KGs, consequently leading to poor data analytics and decision making.

- **Healthcare KG reasoning:** Reasoning of the KG aims to infer new facts and make new conclusions based on the existing data. KG reasoning allows for deriving new insights and enriches KGs with new relations. Several techniques have been proposed in the literature for KG reasoning, including ontology reasoning, logic rules, and random walk algorithm [161]. Recently, KG embedding approaches attracted a lot of attention in the research community due to their capacity to provide generalizations and infer new facts. KG embedding techniques aim to transform the KG into semantically-continuous low-dimensional space. The embedded KG can be then used for several downstream tasks including link prediction, knowledge discovery, etc. [162]. This study reveals a relative lack of successful KG embedding strategies in the investigated papers.

## 7. Conclusion

The vast volume of healthcare data that is collected in a variety of formats and pertains to a wide range of subject matters presents a critical challenge for analysts. Knowledge Graphs (KGs) offer an effective answer to this challenge and open new possibilities for machines to understand meanings, closing the semantic gap between them and people. As a result, domain-specific knowledge graphs have been developed and applied to various real-world problems. Healthcare industry has greatly benefited from this technology, with numerous KGs created specifically to address different healthcare issues. However, the deficiencies and limitations of the current KG construction techniques stand in the way of obtaining the hoped-for value from this technology.

This paper offers a bird's eye view of the healthcare KG domain and tries to define a relevant construction paradigm. A critical review of the current construction approaches is conducted considering the methods used for knowledge extraction, types of knowledge bases and sources, and the adopted evaluation metrics. Finally, in conjunction with a summary of limitations and deficiencies, it provides pointers for potential future research that we hope will inspire scholars in this field.

## 8. Declarations

### Ethics approval and consent to participate

Not applicable.

### Consent for publication

We give the publisher the permission of the authors to publish the work.

### Availability of data and materials

Not applicable.

### Competing interests

The authors declare that they have no known competing financial interests or personal relationships that could have appeared to influence the work reported in this paper.

### Funding

Not applicable.

### Authors' contributions

BAS: Conceptualization, Methodology, Visualization, Writing original draft. MQ: Writing, Review & Editing. MAW and MAS: Validation, Review & Editing. RAF and HS: Supervision, Review & Editing.

### Acknowledgements

Not applicable.